\documentclass[11pt,letterpaper]{article}
\usepackage{ijcnlp2017}
\usepackage{times}
\usepackage{latexsym}
\usepackage{graphicx}

\ijcnlpfinalcopy

\def\ijcnlppaperid{***}

\newcommand\blfootnote[1]{%
  \begingroup
  \renewcommand\thefootnote{}\footnote{#1}%
  \addtocounter{footnote}{-1}%
  \endgroup
}

\title{Text Summarization using \\ Abstract Meaning Representation}

\author{Shibhansh Dohare \\
  CSE Department, IIT Kanpur \\
  {\tt sdohare@cse.iitk.ac.in} \\
  \And
  Harish Karnick \\
  CSE Department, IIT Kanpur\\
  {\tt hk@cse.iitk.ac.in} \\
  \And
  Vivek Gupta \\
  Microsoft Research, Bangalore\\
  {\tt t-vigu@microsoft.com}\\}

\date{}

\begin{document}

\maketitle

\begin{abstract}

With an ever increasing size of text present on the Internet, automatic summary generation remains an important problem for natural language understanding. In this work we explore a novel full-fledged pipeline for text summarization with an intermediate step of Abstract Meaning Representation (AMR). The pipeline proposed by us first generates an AMR graph of an input story, through which it extracts a summary graph and finally, generate summary sentences from this summary graph. Our proposed method achieves state-of-the-art results compared to the other text summarization routines based on AMR. We also point out some significant problems in the existing evaluation methods, which make them unsuitable for evaluating summary quality.

\end{abstract}

\section{Introduction}
Summarization of large texts is still an open problem in language processing. People nowadays have lesser time and patience to go through large pieces of text which make automatic summarization important. Automatic summarization has significant applications in summarizing large texts like stories, journal papers, news articles and even larger texts like books.

Existing methods for summarization can be broadly categorized into two categories \textit{Extractive} and \textit{Abstractive}. Extractive methods picks up words and sometimes directly sentences from the text. These methods are inherently limited in the sense that they can never generate human level summaries for large and complicated documents which require rephrasing sentences and incorporating information from full text to generate summaries. Most of the work done on summarization in past has been extractive.

On the other hand most Abstractive methods take advantages of the recent developments in deep learning. Specifically the recent success of the sequence to sequence learning models where recurrent networks read the text, encodes it and then generate target text. Though these methods have recently shown to be competitive with the extractive methods they are still far away from reaching human level quality in summary generation.

The work on summarization using AMR was started by ~\citet{AMR_Summarization}. Abstract Meaning Representation (AMR) was as introduced by ~\citet{AMR}. AMR focuses on capturing the meaning of the text, by giving a specific meaning representation to the text. AMR tries to capture the \textit{"who is doing what to whom"} in a sentence. The formalism aims to give same representation to sentences which have the same underlying meaning. For example \textit{"He likes apple"} and \textit{"Apples are liked by him"} should be assigned the same AMR.

~\citet{AMR_Summarization}'s approach aimed to produce a summary for a story by extracting a summary subgraph from the story graph and finally generate a summary from this extracted graph. But, because of the unavailability of AMR to text generator at that time their work was limited till extracting the summary graph. This method extracts a single summary graph from the story graph. Extracting a single summary graph assumes that all of the important information from the graph can be extracted from a single subgraph. But, it can be difficult in cases where the information is spread out in the graph. Thus, the method compromises between size of the summary sub-graph and the amount of information it can extract. This can be easily solved if instead of a single sub-graph, we extract multiple subgraphs each focusing on information in a different part of the story.

We propose a two step process for extracting multiple summary graphs. First step is to select few sentences from the story. We use the idea that there are only few sentences that are important from the point of view of summary, i.e. most of the information contained in the summary is present in very few sentences and they can be used to generate the summary. Second step is to extract important information from the selected sentences by extracting a sub-graph from the selected sentences.

Our main contributions in this work are three folds, 
\begin{itemize}
\item We propose a full-fledged pipeline for text summarization, providing strong baseline for future work on summarization using AMR.

\item Present a novel approach for extracting multiple summary graphs that outperforms the previous methods based on a single sub-graph extraction.

\item Expose some problems with existing evaluation methods and datasets for abstractive summarization.
\end{itemize}

The rest of the paper is organized as follows. Section~\ref{sec:2} contains introduction to AMR, section~\ref{sec:3} and~\ref{sec:4} contains the datasets and the algorithm used for summary generation respectively. Section~\ref{sec:5} has a detailed step-by-step evaluation of the pipeline and in section~\ref{sec:6} we discuss the problems with the current dataset and evaluation metric.

\begin{figure}
\centering
\includegraphics[scale=0.4,keepaspectratio]{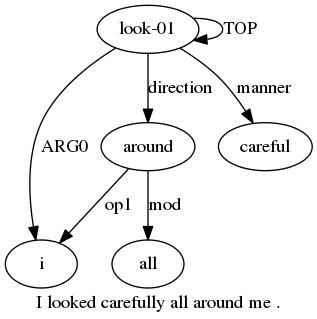}
\label{exmfig}
\caption{The graphical representation of the AMR graph of the sentence : \textit{"I looked carefully all around me"} using AMRICA}
\label{fig:1}
\end{figure}

\section{Background: AMR Parsing and Generation}
\label{sec:2}

AMR was introduced by ~\citet{AMR} with the aim to induce work on statistical Natural Language Understanding and Generation. AMR represents meaning using graphs. AMR graphs are rooted, directed, edge and vertex labeled graphs. Figure~\ref{fig:1} shows the graphical representation of the AMR graph of the sentence \textit{"I looked carefully all around me"} generated by JAMR parser (~\citet{JAMR}). The graphical representation was produced using AMRICA ~\citet{AMRICA}. The nodes in the AMR are labeled with \textit{concepts} as in Figure~\ref{fig:1} \textit{around} represents a concept. Edges contains the information regarding the \textit{relations} between the concepts. In Figure~\ref{fig:1} \textit{direction} is the relation between the concepts \textit{look-01} and \textit{around}. AMR relies on Propbank for semantic relations (edge labels). Concepts can also be of the form \textit{run-01} where the index \textit{01} represents the first sense of the word \textit{run}. Further details about the AMR can be found in the AMR guidelines ~\citet{AMR-Guidelines}.

A lot of work has been done on parsing sentences to their AMR graphs. There are three main approaches to parsing. There is alignment based parsing ~\citet{JAMR} (JAMR-Parser), ~\citet{ZJAMR} which uses graph based algorithms for concept and relation identification. Second, grammar based parsers like ~\citet{CAMR} (CAMR) generate output by performing shift reduce transformations on output of a dependency parser. Neural parsing ~\citet{Neural_AMR,data_sparsity} is based on using \textit{seq2seq} models for parsing, the main problem for neural methods is the absence of a huge corpus of human generated AMRs. ~\citet{data_sparsity} reduced the vocabulary size to tackle this problem while ~\citet{Neural_AMR} used larger external corpus of external sentences.

Recently, some work has been done on producing meaningful sentences form AMR graphs. ~\citet{JAMR} used a number of tree to string conversion rules for generating sentences. ~\citet{song} reformed the problem as a traveling salesman problem. ~\citet{Neural_AMR} used \textit{seq2seq} learning methods.

\section{Datasets}
\label{sec:3}

We used two datasets for the task - AMR Bank ~\citet{AMR_Bank} and CNN-Dailymail (~\cite{cnn_dm} ~\cite{Abstractive_Text_Summarization}). We use the proxy report section of the AMR Bank, as it is the only one that is relevant for the task because it contains the gold-standard (human generated) AMR graphs for news articles, and the summaries. In the training set the stories and summaries contain 17.5 sentences and 1.5 sentences on an average respectively. The training and test sets contain 298 and 33 summary document pairs respectively.

CNN-Dailymail corpus is better suited for summarization as the average summary size is around 3 or 4 sentences. This dataset has around 300k document summary pairs with stories having 39 sentences on average. The dataset comes in 2 versions, one is the anonymized version, which has been preprocessed to replace named entities, e.g., \textit{The Times of India}, with a unique identifier for example \textit{@entity1}. Second is the non-anonymized which has the original text. We use the non-anonymized version of the dataset as it is more suitable for AMR parsing as most of the parsers have been trained on non-anonymized text. The dataset does not have gold-standard AMR graphs. We use automatic parsers to get the AMR graphs but they are not gold-standard and will effect the quality of final summary. To get an idea of the error introduced by using automatic parsers, we compare the results after using gold-standard and automatically generated AMR graphs on the gold-standard dataset.

\section{Pipeline for Summary Generation}
\label{sec:4}
The pipeline consists of three steps, first convert all the given story sentences to there AMR graphs followed by extracting summary graphs from the story sentence graphs and finally generating sentences from these extracted summary graphs. In the following subsections we explain each of the methods in greater detail. 

\subsection{Step 1: Story to AMR}
As the first step we convert the story sentences to their Abstract Meaning Representations. We use JAMR-Parser version 2 \citet{JAMR} as it’s openly available and has a performance close to the state of the art parsers for parsing the CNN-Dailymail corpus. For the AMR-bank we have the gold-standard AMR parses but we still parse the input stories with JAMR-Parser to study the effect of using graphs produced by JAMR-Parser instead of the gold-standard AMR graphs.

\subsection{Step 2: Story AMR to Summary AMR}
After parsing (Step 1) we have the AMR graphs for the story sentences. In this step we extract the AMR graphs of the summary sentences using story sentence AMRs. We divide this task in two parts. First is finding the important sentences from the story and then extracting the key information from those sentences using their AMR graphs.

\subsubsection{Selecting Important Sentences}
\label{ssec:impsnt}
\begin{figure}
\includegraphics[scale=0.4,keepaspectratio]{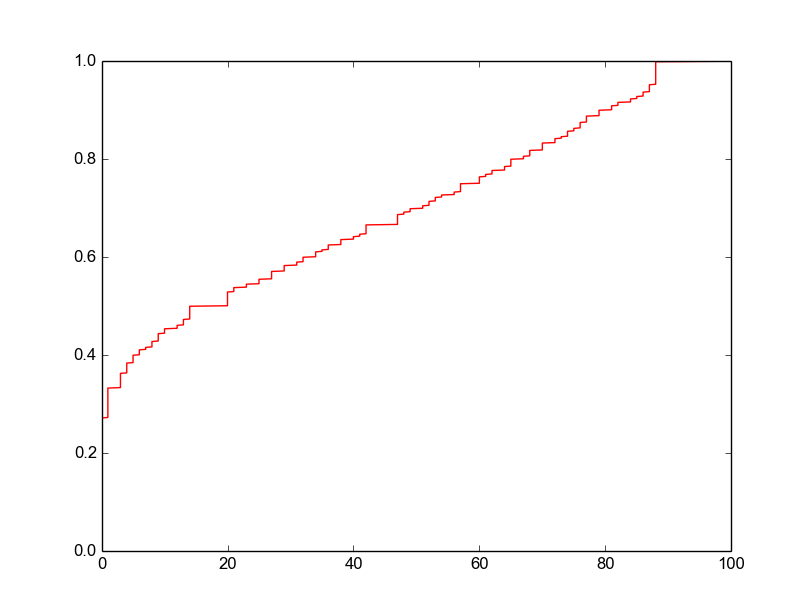}
\caption{\label{graph}Graph of the best Rogue-1 recall scores for 5000 summaries (around 20000 sentences) in the CNN-Dailymail corpus. Y-axis is the ROGUE score and X-axis is the cumulative percentage of sentence with the corresponding score}
\end{figure}

Our algorithm is based on the idea that only few sentences are important from the point of view of summary i.e. there are only a few sentences which contain most of the important information and from these sentences we can generate the summary.

\textbf{Hypothesis:} Most of the information corresponding to a summary sentence can be found in only one sentence from the story.

To test this hypothesis, for each summary sentence we find the sentence from the story that contains maximum information of this summary sentence. We use ROGUE-1 ~\citet{ROGUE} Recall scores (measures the ratio of number of words in the target summary that are contained in the predicted summary to the total number of words in the target summary) as the metric for the information contained in the story sentence. We consider the story sentence as the predicted summary and the summary sentence as the target summary. The results that we obtained for 5000 randomly chosen document summary pairs from the CNN-Dailymail corpus are given in figure~\ref{graph}. The average recall score that we obtained is 79\%. The score will be perfectly 1 when the summary sentence is directly picked up from a story sentence. Upon manual inspection of the summary sentence and the corresponding best sentence from the story we realized, when this score is more than 0.5 or 0.6, almost always the information in the summary sentence is contained in this chosen story sentence. The score for in these cases is not perfectly 1 because of stop words and different verb forms used in story and summary sentence. Around 80\% of summary sentences have score above 0.5. So, our hypothesis seems to be correct for most of the summary sentences. This also suggests the highly extractive nature of the summary in the corpus.

Now the task in hand is to select few important sentences. Methods that use sentence extraction for summary generation can be used for the task. It is very common in summarization tasks specifically in news articles that a lot of information is contained in the initial few sentences. Choosing initial few sentences as the summary produces very strong baselines which the state-of-the-art methods beat only marginally. Ex. On the CNN-Dailymail corpus the state-of-the-art extractive method beats initial 3 sentences only by 0.4\% as reported by ~\cite{summarunner}.

\blfootnote{Note that methods \textit{first-n} and \textit{first co-occurrence+first} are by default followed by the summary graph extraction step and they are not just sentence selection methods.}

Using this idea of picking important sentences from the beginning, we propose two methods, first is to simply pick initial few sentences, we call this \textit{first-n} method where n stands for the number of sentences. We pick initial 3 sentences for the CNN-Dailymail corpus i.e. \textit{first-3} and only the first sentence for the proxy report section (AMR Bank) i.e. \textit{first-1} as they produce the best scores on the ROGUE metric compared to any other \textit{first-n}. Second, we try to capture the relation between the two most important entities (we define importance by the number of occurrences of the entity in the story) of the document. For this we simply find the first sentence which contains both these entities. We call this the \textit{first co-occurrence} based sentence selection. We also select the first sentence along with \textit{first co-occurrence} based sentence selection as the important sentences. We call this the \textit{first co-occurrence+first} based sentence selection.

\subsubsection{Extracting Summary Graph}
\label{ssec:extract-AMR}
As the datasets under consideration are news articles. The most important information in them is about an entity and a verb associated with it. So, to extract important information from the sentence. We try to find the entity being talked about in the sentence, we consider the most referred entity (one that occurs most frequently in the text), now for the main verb associated with the entity in the sentence, we find the verb closest to this entity in the AMR graph. We define the closest verb as the one which lies first in the path from the entity to the root.

We start by finding the position of the most referred entity in the graph, then we find the closest verb to the entity. and finally select the subtree hanging from that verb as the summary AMR.

\subsection{Step 3: Summary Generation}
To generate sentences from the extracted AMR graphs we can use already available generators. We use Neural AMR (~\citet{Neural_AMR}) as it provides state of the art results in sentence generation. We also use ~\citet{JAMR2} (JAMR-Generator) in one of the experiments in the next section. Generators significantly effect the results, we will analyze the effectiveness of generator in the next section. 

\section{Results and Analysis}
\label{sec:5}
\subsection{Baselines}
\label{ssec:baselines}
\label{ssec:evaluation}

\begin{table*}[h]
\centering
\caption{\label{proxy-section-results}Comparison with previous methods and baselines. This table reports ROGUE scores on the proxy report section using alignment based generator.}
\vspace{1.5mm}
	\begin{tabular}{c|c|c|c}
      \hline
	Method & Rogue-1 Recall & Rogue-1 Precision & Rogue-1 $F_{1}$\\
      \hline
      	~\citet{AMR_Summarization} & 51.9 & 39.0 & 44.3 \\
	    Lead-1-AMR & 50.4 & 57.5 & 51.0 \\
		fist co-occurrence + first & ~\textbf{52.4} & 55.7 & \textbf{51.3}\\
		first-1 & 49.1 & \textbf{60.1} & 51.2\\
\hline
    \end{tabular}
  \end{table*}

\begin{table*}[h]
\centering
    \caption{\label{final-results}Table for analyzing the effect of using JAMR-Parser in step-1. This table has ROGUE scores after using Neural AMR for sentence generation i.e. step-3. First half contains scores by using gold-standard AMR graphs, second half has AMR graphs generated by JAMR-Parser}
\vspace{2.0mm}
	\begin{tabular}{c|c|c|c|c|c}
      \hline
	Method & Rogue-1 Recall & Rogue-1 Precision & Rogue-1 $F_{1}$ & Rogue-2 & Rogue-L\\
      \hline
    \multicolumn{6}{c}{Using gold-standard AMR in step-1}\\
    \cline{1-6}
	Lead-1-AMR & 46.8 & 49.0 & 45.5 & 21.5 & \textbf{35.2} \\
	first co-occurrence + first & \textbf{49.5} & 48.1 & \textbf{46.3} & 21.7 & 34.7 \\
	first-1 & 45.9 & \textbf{51.4} & 45.9 & \textbf{21.9} & 35.0 \\
	\cline{1-6}
\multicolumn{6}{c}{Using JAMR-Parser for step-1}\\
\cline{1-6}
	Lead-1-AMR & 43.7 & 44.7 & \textbf{41.4} & 16.2 & \textbf{28.3} \\
	first co-occurrence + first  & \textbf{44.5} & 42.4 & 40.0 & \textbf{17.0} & 27.5 \\
	first-1  & 41.1 & \textbf{45.4} & 40.1 & 15.3 & 28.2 \\
	\cline{1-6}
\hline
    \end{tabular}
  \end{table*}

In this section, we present the baseline models and analysis method used for each step of our pipeline.

For the CNN-Dailymail dataset, the Lead-3 model is considered a strong baseline; both the abstractive ~\cite{deep_rl} and extractive ~\cite{summarunner} state-of-the art methods on this dataset beat this baseline only marginally. The Lead-3 model simply produces the leading three sentences of the document as its summary. 

The key step in our pipeline is step-2 i.e. summary graph extraction. 

Directly comparing the Lead-3 baseline, with AMR based pipeline to evaluate the effectiveness of step-2 is an unfair comparison because of the errors introduced by imperfect parser and generator in the AMR pipeline. Thus to evaluate the effectiveness of step-2 against Lead-3 baseline, we need to nullify the effect of errors introduce by AMR parser and generator. We achieve this by trying to introduce similar errors in the leading thre sentences of each document. We generate the AMR graphs of the leading three sentences and then generate the sentences using these AMR graph. We use parser and generator that were used in our pipeline. We consider these generated sentences as the new baseline summary, we shall now refer to it \textit{Lead-3-AMR} baseline in the remaining of the paper.

For the proxy report section of the AMR bank, we consider the \textit{Lead-1-AMR} model as the baseline. For this dataset we already have the gold-standard AMR graphs of the sentences. Therefore, we only need to nullify the error introduced by the generator.

\subsection{Procedure to Analyze and Evaluate each step}

For the evaluation of summaries we use the standard ROGUE metric. For comparison with previous AMR based summarization methods, we report the \textit{Recall}, \textit{Precision} and \textit{$F_{1}$} scores for ROGUE-1. Since most of the literature on summarization uses \textit{$F_{1}$} scores for ROGUE-2 and ROGUE-L for comparison, we also report \textit{$F_{1}$} scores for ROGUE-2 and ROGUE-L for our method. ROGUE-1 Recall and Precision are measured for uni-gram overlap between the reference and the predicted summary. On the other hand, ROGUE-2 uses bi-gram overlap while ROGUE-L uses the longest common sequence between the target and the predicted summaries for evaluation. In rest of this section, we provide methods to analyze and evaluate our pipeline at each step.

\textbf{Step-1: AMR parsing} To understand the effects of using an AMR parser on the results, we compare the final scores after the following two cases- first, when we use the gold-standard AMR graphs and second when we used the AMR graphs generated by JAMR-Parser in the pipeline. Section~\ref{ssec:jamr_effect} contains a comparison between the two.

\textbf{Step-2: Summary graph extraction} For evaluating the effectiveness of the summary graph extraction step we compare the final scores with the \textit{Lead-n-AMR} baselines described in section~\ref{ssec:baselines}.

In order to compare our summary graph extraction step with the previous work (~\citet{AMR_Summarization}), we generate the final summary using the same generation method as used by them. Their method uses a simple module based on alignments for generating summary after step-2. The alignments simply map the words in the original sentence with the node or edge in the AMR graph. To generate the summary we find the words aligned with the sentence in the selected graph and output them in no particular order as the predicted summary. Though this does not generate grammatically correct sentences, we can still use the ROGUE-1 metric similar to ~\citet{AMR_Summarization}, as it is based on comparing uni-grams between the target and predicted summaries.

\textbf{Step-3: Generation} For evaluating the quality of the sentences generated by our method, we compare the summaries generated by the first-1 model and Lead-1-AMR model on the gold-standard dataset. However, when we looked at the scores given by ROGUE, we decided to do get the above summaries evaluated by humans. This produced interesting results which are given in more detail in section~\ref{ssec:generator_effect}.

\subsection{Results on the Proxy report section}
\begin{table*}[h]
\centering
\caption{\label{human_evaluation}Comparion of the scores given by ROGUE and human evaluators on different models. Scores suggest that Rogue don't co-relate with the human evaluators}
\vspace{1.5mm}
	\begin{tabular}{c|c|c|c|c|c|c}
      \hline
	Parser & Generator & Information contained & Readability & R-1 & R-2 & R-L \\
\hline
      	gold & jamr & 5.24 & 5.48 & 43.3 & 16.3 & 29.9 \\
	    jamr & neural & 4.52 & 7.04 & 47.7 & 19.0 & 32.9 \\
		gold & neural & 6.19 & \textbf{7.88} & 49.7 & 25.0 & 38.0 \\
        \hline
\multicolumn{2}{c|}{original sentence} & \textbf{6.88} & 7.4 & \textbf{62.6} & \textbf{48.6} & \textbf{54.7}\\
	\hline
    \end{tabular}
\end{table*}
\begin{table*}[h]
\centering
\caption{\label{fig:CNN-Results}Results on CNN-Dailymail corpus. Table has 2 parts. First part contains baselines, our method and the state-of-the-art on the non-anonymized dataset, second part has scores on the anonymized dataset.}
\vspace{1.5mm}
	\begin{tabular}{c|c|c|c|c|c}
      \hline
Method & \multicolumn{5}{c}{ROGUE} \\
\cline{2-6}
  & 1 Recall & 1 Precision & 1 & 2 & L  \\
      \hline
Lead-3-AMR (baseline) & 40.4 &  27.8 & 31.7 & 5.8 & 16.8 \\
first-3 & 38.1 &  28.8 &	31.6 & 5.7 & 16.9 \\
Lead-3 (non-anonymized) ~\cite{pointer-generator} &-&- &40.34 &17.70 &36.57\\
pointer-generator + coverage ~\cite{pointer-generator} &- &- &39.53 &17.28&36.38 \\
      \hline
Lead-3 (anonymized) ~\cite{summarunner} &- &- &39.2 &15.7 &35.5 \\
RL, with intra-attention ~\cite{deep_rl} &-&- &41.16 &15.75 &39.08 \\
\hline
\end{tabular}
\end{table*}

In table~\ref{proxy-section-results} we report the results of using the pipeline with generation using the alignment based generation module defined in section~\ref{ssec:evaluation}, on the proxy report section of the AMR Bank. All of our methods  out-perform ~\citet{AMR_Summarization}'s method. We obtain best ROGUE-1 $F_{1}$ scores using the ~\textit{first co-occurrence+first} model for important sentences. This also out-perform our ~\textit{Lead-1-AMR} baseline by 0.3 ROGUE-1 $F_{1}$ points.

\subsection{Effects of using JAMR Parser}
\label{ssec:jamr_effect}
In this subsection we analyze the effect of using JAMR parser for step-1 instead of the gold-standard AMR graphs. First part of table~\ref{final-results} has scores after using the gold-standard AMR graphs. In the second part of table~\ref{final-results} we have included the scores of using the JAMR parser for AMR graph generation. We have used the same Neural AMR for sentence generation in all methods. Scores of all methods including the ~\textit{Lead-1-AMR} baseline have dropped significantly. 

The usage of JAMR Parser has affected the scores of \textit{first co-occurrence+first} and \textit{first-1} more than that for the \textit{Lead-1-AMR}. The drop in ROGUE $F_{1}$ score when we use \textit{first co-occurrence+first} is around two ROGUE $F_{1}$ points more than when \textit{Lead-1-AMR}. This is a surprising result, and we believe that it is worthy of further research.

\subsection{Effectiveness of the Generator}
\label{ssec:generator_effect}

In this subsection we evaluate the effectiveness of the sentence generation step. For fair comparison at the generation step we use the gold-standard AMRs and don't perform any extraction in step-2 instead we use full AMRs, this allows to remove any errors that might have been generated in step-1 and step-2. In order to compare the quality of sentences generated by the AMR, we need a gold-standard for sentence generation step. For this, we simply use the original sentence as gold-standard for sentence generation. Thus, we compare the quality of summary generated by \textit{Lead-1} and \textit{Lead-1-AMR}. The scores using the ROGUE metric are given in bottom two rows of table~\ref{human_evaluation}. The results show that there is significant drop in \textit{Lead-1-AMR} when compared to \textit{Lead-1}.

We perform human evaluation to check whether the drop in ROGUE scores is because of drop in information contained, and human readability or is it because of the inability of the ROGUE metric to judge. To perform this evaluation we randomly select ten test examples from the thirty- three test cases of the proxy report section. For each example, we show the summaries generated by four different models side by side to the human evaluators. The human evaluator does not know which summaries come from which model. A score from 1 to 10 is then assigned to each summary on the basis of \textit{readability}, and \textit{information contained} of summary, where 1 corresponds to the lower level and 10 to the highest. In table~\ref{human_evaluation} we compare the scores of these four cases as given by ROGUE along with human evaluation. The parser-generator pairs for the four cases are  gold-JAMR(generator), JAMR(parser)-neural, gold-neural, and the original sentence respectively. Here gold parser means that we have used the gold-standard AMR graphs.

The scores given by the humans do not correlate with ROGUE. Human evaluators gives almost similarly scores to summary  generated by the \textit{Lead-1} and \textit{Lead-1-AMR} with \textit{Lead-1-AMR} actually performing better on \textit{readability} though it dropped some information as clear from the scores on \textit{information contained}. On the other hand, ROGUE gives very high score to Lead-1 while models 1,2 and 4 get almost same scores. The similar scores of model 2 and 3 shows that generators are actually producing meaningful sentences. Thus the drop in ROGUE scores is mainly due to the inability of the ROGUE to evaluate abstractive summaries. Moreover, the ROGUE gives model 4 higher score compared to model 1 while human evaluators give the opposite scores on ~\textit{information contained} in the sentence.

A possible reason for the inability of the ROGUE metric to properly evaluate the summaries generated by our method might be due to its inability to evaluate restructured sentences. AMR formalism tries to assign the same AMR graphs to the sentences that have same meaning so there exists a one-to-many mapping from AMR graphs to sentences. This means that the automatic generators that we are using might not be trying to generate the original sentence; instead it is trying to generate some other sentence that has the same underlying meaning. This also helps in explaining the low ROGUE-2 and ROGUE-L scores. If the sentences might be getting rephrased, they would loose most of the bi- and tri-grams from the original sentence resulting in low ROGUE-2 and ROGUE-L scores.

\subsection{Analyzing the effectiveness of AMR extraction}

The aim of extracting summary graphs from the AMR graphs of the sentence is to drop the not so important information from the sentences. If we are able to achieve this perfectly, the ROGUE-1 Recall scores that we are getting should remain almost the same (since we are not add any new information) and the ROGUE-1 precision should go up (as we have thrown out some useless information); thus effectively improving the overall ROGUE-1 $F_{1}$ score. In the first two rows of table ~\ref{final-results} we have the scores after using the full-AMR and extracted AMR for generation respectively. It is safe to say that extracting the AMR results in improved ROGUE-1 precision whereas ROGUE-1 Recall reduces only slightly, resulting in an overall improved ROGUE-1 $F_{1}$.

\subsection{Results on the CNN-Dailymail corpus}

In table~\ref{fig:CNN-Results} we report the results on the CNN-Dailymail corpus. We present scores by using the \textit{first-3} model. The first row contains the \textit{Lead-3-AMR} baseline. The results we achieve are competitive with the \textit{Lead-3-AMR} baseline. The rest of the table contains scores of Lead-3 baseline followed by the state-of-the-art method on the anonymized and non-anonymized versions of the dataset. The drop in the scores from the Lead-3(non-anonymized) to \textit{Lead-3-AMR} is significant and is largely because of the error introduced by parser and generator. 

\section{Related Work and Discussion}
\label{sec:6}
\subsection{Related Work}

~\citet{tac2008} showed that most of the work in text summarization has been extractive, where sentences are selected from the text which are then concatenated to form a summary. ~\citet{Vanderwende2004} transformed the input to nodes, then used the Pagerank algorithm to score nodes, and finally grow the nodes from high-value to low-value using some heuristics. Some of the approaches combine this with sentence compression, so more sentences can be packed in the summary. ~\citet{raynmcd2007}, ~\citet{Martins2009}, ~\citet{Almeida2013}, and ~\citet{Gillick2009} among others used ILPs and approximations for encoding compression and extraction.

Recently some abstractive approaches have also been proposed most of which used sequence to sequence learning models for the task. ~\cite{neural_attention}, ~\cite{neural_attention_sentence}, ~\cite{Abstractive_Text_Summarization}, ~\cite{pointer-generator} used standard encoder-decoder models along with their variants to generate summaries. ~\cite{NeuralheadlineAMR} incorporated the AMR information in the standard encoder-decoder models to improve results. Our work in similar to other graph based abstractive summarization methods ~\cite{Penn2014} and ~\citet{Gerani2014}. ~\cite{Penn2014} used dependency parse trees to produce summaries. On the other hand our work takes advantage of semantic graphs. 

\subsection{Need of an new Dataset and Evaluation Metric}
\label{ssec:rogue_problems}

ROGUE metric, by it is design has lots of properties that make it unsuitable for evaluating abstractive summaries. For example, ROGUE matches exact words and not the stems of the words, it also considers stop words for evaluation. One of the reasons why ROGUE like metrics might never become suitable for evaluating abstractive summaries is its incapabilities of knowing if the sentences have been restructured. A good evaluation metric should be one where we compare the meaning of the sentence and not the exact words. As we showed section ~\ref{ssec:generator_effect} ROGUE is not suitable for evaluating summaries generated by the AMR pipeline.

We now show why the CNN-Dailymail corpus is not suitable for Abstractive summarization. The nature of summary points in the corpus is highly extractive (Section ~\ref{ssec:impsnt} for details) where most of the summary points are simply picked up from some sentences in the story. Tough, this is a good enough reason to start searching for better dataset, it is not the biggest problem with the dataset. The dataset has the property that a lot of important information is in the first few sentences and most of the summary points are directly pick from these sentences. The extractive methods based on sentence selection like SummaRunNer are not actually performing well, the results they have got are only slightly better than the Lead-3 baseline. The work doesn't show how much of the selected sentences are among the first few and it might be the case that the sentences selected by the extractive methods are mostly among the first few sentences, the same can be the problem with the abstractive methods, where most the output might be getting copied from the initial few sentences. 

These problems with this corpus evoke the need to have another corpus where we don't have so much concentration of important information at any location but rather the information is more spread out and the summaries are more abstractive in nature.

\section{Possible Future Directions}
As this proposed algorithm is a step by step process we can focus on improving each step to produce better results. The most exciting improvements can be done in the summary graph extraction method. Not a lot of work has been done to extract AMR graphs for summaries. In order to make this pipeline generalizable for any sort of text, we need to get rid of the hypothesis that the summary is being extracted exactly from one sentence. So, the natural direction seems to be joining AMR graphs of multiple sentences that are similar and then extracting the summary AMR from that large graph. It will be like clustering similar sentences and then extracting a summary graph from each of these cluster. Another idea is to use AMR graphs for important sentence selection.

\section{Conclusion}

In this work we have explored a full-fledged pipeline using AMR for summarization for the first time. We propose a new method for extracting summary graph, which outperformed previous methods. Overall we provide strong baseline for text summarization using AMR for possible future works. We also showed that ROGUE can't be used for evaluating the abstractive summaries generated by our AMR pipeline.



\bibliography{ijcnlp2017}

\begin{thebibliography}{}
\expandafter\ifx\csname natexlab\endcsname\relax\def\natexlab#1{#1}\fi

\bibitem[{Almeida and Martins(2013)}]{Almeida2013}
Miguel~B. Almeida and Andre F.~T. Martins. 2013.
\newblock Fast and robust compressive summarization with dual de-composition
  and multi-task learning.
\newblock In Proceedings of ACL.

\bibitem[{Banarescu et~al.(2015)Banarescu, Bonial, Cai, Georgescu, Griffitt,
  Hermjakob, Knight, Koehn, Palmer, and Schneider}]{AMR-Guidelines}
Laura Banarescu, Claire Bonial, Shu Cai, Madalina Georgescu, Kira Griffitt, Ulf
  Hermjakob, Kevin Knight, Philipp Koehn, Martha Palmer, and Nathan Schneider.
  2015.
\newblock
  \href{https://github.com/amrisi/amr-guidelines/blob/master/amr.md}{Amr-guidelines}.
\newblock
  \href{https://github.com/amrisi/amr-guidelines/blob/master/amr.md}{https://github.com/amrisi/amr-guidelines/blob/master/amr.md}.

\bibitem[{Banarescu et~al.(2013)Banarescu, Bonial, Cai, Georgescu, Griffitt,
  Hermjakob, Knight, Palmer, Koehn, and Schneider}]{AMR}
Laura Banarescu, Claire Bonial, Shu Cai, Madalina Georgescu, Kira Griffitt, Ulf
  Hermjakob, Kevin Knight, Martha Palmer, Philipp Koehn, and Nathan Schneider.
  2013.
\newblock \href{http://www.aclweb.org/anthology/W13-2322}{Abstract meaning
  representation for sembanking}.
\newblock Proceedings of Linguistic Annotation Workshop.
\newblock
  \href{http://www.aclweb.org/anthology/W13-2322}{http://www.aclweb.org/anthology/W13-2322}.

\bibitem[{Chopra et~al.(2016)Chopra, Auli, and
  Rush}]{neural_attention_sentence}
Sumit Chopra, Michael Auli, and Alexander~M. Rush. 2016.
\newblock Abstractive sentence summarization with attentive recurrent neural
  networks.

\bibitem[{Dang and Owczarzak(2008)}]{tac2008}
Hoa~Trang Dang and Karolina Owczarzak. 2008.
\newblock Overview of the tac 2008 update summarization task.
\newblock In Proceedings of Text Analysis Conference (TAC).

\bibitem[{Flanigan et~al.(2016)Flanigan, Carbonell, Dyer, and Smith}]{JAMR2}
Jeffrey Flanigan, Jaime Carbonell, Chris Dyer, and Noah~A. Smith. 2016.
\newblock \href{https://aclweb.org/anthology/D16-1065.}{Generation from
  abstract meaning representation using tree transducers}.
\newblock In Proceedings of the 2016 Conference of the North American Chapter
  of the Association for Computational Linguistics.
\newblock
  \href{https://aclweb.org/anthology/D16-1065.}{https://aclweb.org/anthology/D16-1065.}

\bibitem[{Flanigan et~al.(2014)Flanigan, Thomson, Carbonell, Dyer, and
  Smith}]{JAMR}
Jeffrey Flanigan, Sam Thomson, Jaime Carbonell, Chris Dyer, and Noah~A. Smith.
  2014.
\newblock \href{http://www.aclweb.org/anthology/P14-1134}{A discriminative
  graph-based parser for the abstract meaning representation}.
\newblock In Proceedings of the 52nd Annual Meeting of the Association for
  Computational Linguistics. Association for Computational Linguistics,
  Baltimore, Maryland, pages 1426–1436.
\newblock
  \href{http://www.aclweb.org/anthology/P14-1134}{http://www.aclweb.org/anthology/P14-1134}.

\bibitem[{Gerani et~al.(2014)Gerani, Mehdad, Carenini, Ng, and
  Nejat}]{Gerani2014}
Shima Gerani, Yashar Mehdad, Giuseppe Carenini, Raymond~T. Ng, and Bita Nejat.
  2014.
\newblock \href{http://emnlp2014.org/papers/pdf/EMNLP2014168.pdf}{Abstractive
  summarization of product reviews using discourse structure}.
\newblock In Proceedings of EMNLP.
\newblock
  \href{http://emnlp2014.org/papers/pdf/EMNLP2014168.pdf}{http://emnlp2014.org/papers/pdf/EMNLP2014168.pdf}.

\bibitem[{Gillick and Favre(2009)}]{Gillick2009}
Dan Gillick and Benoit Favre. 2009.
\newblock \href{http://dl.acm.org/citation.cfm?id=1611640}{A scalable global
  model for summarization}.
\newblock In Proceedings of the NAACL Workshop on Integer Linear Programming
  for Natural Langauge Processing.
\newblock
  \href{http://dl.acm.org/citation.cfm?id=1611640}{http://dl.acm.org/citation.cfm?id=1611640}.

\bibitem[{Hermann et~al.(2015)Hermann, Kocisky, Grefenstette, Espeholt, Kay,
  Suleyman, and Blunsom}]{cnn_dm}
Karl~Moritz Hermann, Tomas Kocisky, Edward Grefenstette, Lasse Espeholt, Will
  Kay, Mustafa Suleyman, and Phil Blunsom. 2015.
\newblock \href{https://arxiv.org/pdf/1506.03340.pdf}{Teaching machines to read
  and comprehend}.
\newblock
  \href{https://arxiv.org/pdf/1506.03340.pdf}{https://arxiv.org/pdf/1506.03340.pdf}.

\bibitem[{Knight et~al.(2014)Knight, Baranescu, Bonial, Georgescu, Griffitt,
  Hermjakob, Marcu, Palmer, and Schneider}]{AMR_Bank}
Kevin Knight, Laura Baranescu, Claire Bonial, Madalina Georgescu, Kira
  Griffitt, Ulf Hermjakob, Daniel Marcu, Martha Palmer, and Nathan Schneider.
  2014.
\newblock Deft phase 2 amr annotation r1 ldc2015e86. philadelphia: Linguistic
  data consortium.
\newblock Abstract meaning representation (AMR) annotation release 1.0
  LDC2014T12. Web Download. Philadelphia: Linguistic Data Consortium.

\bibitem[{Konstas et~al.(2017)Konstas, Iyer, Yatskar, Choi, and
  Zettlemoyer}]{Neural_AMR}
Ioannis Konstas, Srinivasan Iyer, Mark Yatskar, Yejin Choi, and Luke
  Zettlemoyer. 2017.
\newblock \href{https://arxiv.org/abs/1704.08381}{Neural amr:
  Sequence-to-sequence models for parsing and generation}.
\newblock In Proceedings of the 52nd Annual Meeting of the Association for
  Computational Linguistics. Association for Computational Linguistics.
\newblock
  \href{https://arxiv.org/abs/1704.08381}{https://arxiv.org/abs/1704.08381}.

\bibitem[{Lin(2004)}]{ROGUE}
C.~Lin. 2004.
\newblock Rouge: A package for automatic evaluation of summaries.
\newblock Text Summarization Branches Out, Post-Conference Workshop of ACL
  2004. Barcelona, Spain.

\bibitem[{Liu et~al.(2015)Liu, Jeffrey, Sam, Norman, and
  Noah~A.}]{AMR_Summarization}
Fei Liu, Flanigan Jeffrey, Thomson Sam, Sadeh Norman, and Smith Noah~A. 2015.
\newblock \href{http://www.aclweb.org/anthology/N15-1114}{Toward abstractive
  summarization using semantic representations}.
\newblock In Proceedings of the 2015 Conference of the North American Chapter
  of the Association for Computational Linguistics. Association for
  Computational Linguistics, Denver, Colorado, pages 1077–1086.
\newblock
  \href{http://www.aclweb.org/anthology/N15-1114}{http://www.aclweb.org/anthology/N15-1114}.

\bibitem[{Martins and Smith(2009)}]{Martins2009}
Andre F.~T. Martins and Noah~A. Smith. 2009.
\newblock \href{http://www.aclweb.org/anthology/W09-1801}{Summarization with a
  joint model for sentence extraction and compression.}
\newblock In Proceedings of the ACL Workshop on Integer Linear Programming for
  Natural Language Processing.
\newblock
  \href{http://www.aclweb.org/anthology/W09-1801}{http://www.aclweb.org/anthology/W09-1801}.

\bibitem[{McDonald(2007)}]{raynmcd2007}
Ryan McDonald. 2007.
\newblock A study of global inference algorithms in multi-document
  summarization.
\newblock In Proceedings of ECIR.

\bibitem[{Nallapati et~al.(2017)Nallapati, Zhai, and Zhou}]{summarunner}
Ramesh Nallapati, Feifei Zhai, and Bowen Zhou. 2017.
\newblock \href{http://aclweb.org/anthology/P16-1001}{Summarunner: A recurrent
  neural network based sequence model for extractive summarization of
  documents}.
\newblock Proceedings of the Thirty-First AAAI Conference on Artificial
  Intelligence (AAAI-17).
\newblock
  \href{http://aclweb.org/anthology/P16-1001}{http://aclweb.org/anthology/P16-1001}.

\bibitem[{Nallapati et~al.(2016)Nallapati, Zhou, Santos, ulçehre, and
  Xiang}]{Abstractive_Text_Summarization}
Ramesh Nallapati, Bowen Zhou, Cicero~dos Santos, Ça ̆glar~G ulçehre, and
  Bing Xiang. 2016.
\newblock \href{http://www.aclweb.org/anthology/K16-1028}{Abstractive text
  summarization using sequence-to-sequence rnns and beyond}.
\newblock In Computational Natural Language Learning.
\newblock
  \href{http://www.aclweb.org/anthology/K16-1028}{http://www.aclweb.org/anthology/K16-1028}.

\bibitem[{Paulus et~al.(2017)Paulus, Xiong, and Socher}]{deep_rl}
Romain Paulus, Caiming Xiong, and Richard Socher. 2017.
\newblock \href{http://arxiv.org/abs/1705.04304}{A deep reinforced model for
  abstractive summarization}.
\newblock
  \href{http://arxiv.org/abs/1705.04304}{http://arxiv.org/abs/1705.04304}.

\bibitem[{Peng et~al.(2017)Peng, Wang, Gildea, and Xue}]{data_sparsity}
Xiaochang Peng, Chuan Wang, Daniel Gildea, and Nianwen Xue. 2017.
\newblock \href{http://www.aclweb.org/anthology/E17-1035}{Addressing the data
  sparsity issue in neural amr parsing}.
\newblock In Proceedings of the 15th Conference of the European Chapter of the
  Association for Computational Linguistics. Association for Computational
  Linguistics, Valencia, Spain, pages 366–375.
\newblock
  \href{http://www.aclweb.org/anthology/E17-1035}{http://www.aclweb.org/anthology/E17-1035}.

\bibitem[{Penn and Cheung(2014)}]{Penn2014}
Gerald Penn and Jackie Chi~Kit Cheung. 2014.
\newblock \href{http://www.aclweb.org/anthology/D14-1085}{Unsupervised sentence
  enhancement for automatic summarization.}
\newblock In Proceedings of EMNLP.
\newblock
  \href{http://www.aclweb.org/anthology/D14-1085}{http://www.aclweb.org/anthology/D14-1085}.

\bibitem[{Rush et~al.(2015)Rush, Chopra, and Weston}]{neural_attention}
Alexander~M. Rush, Sumit Chopra, and Jason Weston. 2015.
\newblock \href{http://arxiv.org/abs/1509.00685}{A neural attention model for
  sentence summarization}.
\newblock
  \href{http://arxiv.org/abs/1509.00685}{http://arxiv.org/abs/1509.00685}.

\bibitem[{Saphra and Lopez(2015)}]{AMRICA}
Naomi Saphra and Adam Lopez. 2015.
\newblock \href{http://www.aclweb.org/anthology/N15-3008}{Amrica: an amr
  inspector for cross-language alignments}.
\newblock System Demonstrations of the 2015 Conference of the North American
  Chapter of the Association for Computational Linguistics.
\newblock
  \href{http://www.aclweb.org/anthology/N15-3008}{http://www.aclweb.org/anthology/N15-3008}.

\bibitem[{See et~al.(2017)See, Liu, and Manning}]{pointer-generator}
Abigail See, Peter~J. Liu, and Christopher~D. Manning. 2017.
\newblock \href{http://aclweb.org/anthology/P16-1001}{Get to the point:
  Summarization with pointer-generator networks}.
\newblock
  \href{http://aclweb.org/anthology/P16-1001}{http://aclweb.org/anthology/P16-1001}.

\bibitem[{Song et~al.(2016)Song, Zhang, Peng, Wang, and Gildea}]{song}
Linfeng Song, Yue Zhang, Xiaochang Peng, Zhiguo Wang, and Daniel Gildea. 2016.
\newblock \href{https://aclweb.org/anthology/D16-1224}{Amr-to-text generation
  as a traveling salesman problem}.
\newblock Proceedings of the 2016 Conference on Empirical Methods in Natural
  Language Processing. Association for Computational Linguistics, Austin,
  Texas, pages 2084–2089.
\newblock
  \href{https://aclweb.org/anthology/D16-1224}{https://aclweb.org/anthology/D16-1224}.

\bibitem[{Takase et~al.(2016)Takase, Suzuki, Okazaki, Hirao, and
  Nagata}]{NeuralheadlineAMR}
Sho Takase, Jun Suzuki, Naoaki Okazaki, Tsutomu Hirao, and Masaaki Nagata.
  2016.
\newblock \href{https://aclweb.org/anthology/D16-1112}{Neural headline
  generation on abstract meaning representation}.
\newblock In Proceedings of EMNLP.
\newblock
  \href{https://aclweb.org/anthology/D16-1112}{https://aclweb.org/anthology/D16-1112}.

\bibitem[{Vanderwende et~al.(2004)Vanderwende, Banko, and
  Menezes}]{Vanderwende2004}
Lucy Vanderwende, Michele Banko, and Arul Menezes. 2004.
\newblock Event-centric summary generation.
\newblock In Proceedings of DUC.

\bibitem[{Wang et~al.(2016)Wang, Pradhan, Pan, Ji, and Xue}]{CAMR}
Chuan Wang, Sameer Pradhan, Xiaoman Pan, Heng Ji, and Nianwen Xue. 2016.
\newblock \href{http://www.aclweb.org/anthology/S16-1181}{Camr at semeval-2016
  task 8: An extended transition-based amr parser}.
\newblock In Proceedings of the 10th International Workshop on Semantic
  Evaluation. Association for Compuational Linguistics.
\newblock
  \href{http://www.aclweb.org/anthology/S16-1181}{http://www.aclweb.org/anthology/S16-1181}.

\bibitem[{Zhou et~al.(2016)Zhou, Xu, Uszkoreit, QU, Li, and Gu}]{ZJAMR}
Junsheng Zhou, Feiyu Xu, Hans Uszkoreit, Weiguang QU, Ran Li, and Yanhui Gu.
  2016.
\newblock \href{https://aclweb.org/anthology/D16-1065}{Amr parsing with an
  incremental joint model}.
\newblock In Proceedings of the 2016 Conference on Empirical Methods in Natural
  Language Processing. Association for Computational Linguistics, Austin,
  Texas, pages 680–689.
\newblock
  \href{https://aclweb.org/anthology/D16-1065}{https://aclweb.org/anthology/D16-1065}.

\end{thebibliography}
\bibliographystyle{ijcnlp2017}

\end{document}